\documentclass[11pt,a4paper]{article}
\usepackage[hyperref]{acl2020}
\usepackage{times}
\usepackage{latexsym}

\usepackage{caption}
\usepackage{subcaption}
\usepackage{graphicx}
\graphicspath{{figures/}}
\usepackage{hyperref}
\usepackage{microtype}

\aclfinalcopy 

\makeatletter
\newcommand{\printfnsymbol}[1]{%
  \textsuperscript{\@fnsymbol{#1}}%
}
\makeatother

\title{Cascading Adaptors to Leverage English Data to Improve Performance of Question Answering for Low-Resource Languages}

\author{Hariom A. Pandya \thanks{\  Equal contribution } \ \thanks{\  Corresponding Authors}\\
  Dharmsinh Desai University \\
  Nadiad-Gujarat(India) \\
  \small{\texttt{pandya.hariom@gmail.com}} \\
  \\\And
  Bhavik Ardeshna \printfnsymbol{1}\printfnsymbol{2}\\
  Dharmsinh Desai University \\
  Nadiad-Gujarat(India) \\
  \small{\texttt{ardeshnabhavik@gmail.com}} \\
  \\\And
  Dr. Brijesh S. Bhatt \\
  Dharmsinh Desai University \\
  Nadiad-Gujarat(India) \\
  \small{\texttt{brij.ce@ddu.ac.in}}
  }

\date{}

\begin{document}
\maketitle
\begin{abstract}
Transformer based architectures have shown notable results on many down streaming tasks including question answering. The availability of data, on the other hand, impedes obtaining legitimate performance for low-resource languages. In this paper, we investigate the applicability of pre-trained multilingual models to improve the performance of question answering in low-resource languages. We tested four combinations of language and task adapters using multilingual transformer architectures on seven languages similar to MLQA dataset. Additionally, we have also proposed zero-shot transfer learning of low-resource question answering using language and task adapters. We observed that stacking the language and the task adapters improves the multilingual transformer models' performance significantly for low-resource languages. 

\end{abstract}

\section{Introduction}
Last few years have seen emergence of transformer based pretrained models like BERT\cite{devlin-etal-2019-bert}, XLNet\cite{NEURIPS2019_dc6a7e65}, T5\cite{JMLR:v21:20-074}, XLM-RoBERTa\cite{conneau-etal-2020-unsupervised} etc. The pretrained models have shown significant improvement in various downstream tasks like question answering, NER, Machine translation and speech recognition when used with word level utilities.\cite{delobelle-etal-2020-robbert,pires-etal-2019-multilingual,pfeiffer-etal-2020-adapterhub,pires-etal-2019-multilingual,pandya2021question,10.1145/3461763,murthy-etal-2019-addressing,inproceedings,baxi-etal-2015-morphological,JMLR:v21:20-074}.

The emergence of multilingual models: mBERT \cite{devlin-etal-2019-bert} and XLM-RoBERTa\cite{conneau-etal-2020-unsupervised} made it possible to leverage English data to improve the performance of low-resource languages. In this paper, we continue to investigate the effectiveness of multilingual pretrained transformer models in improving the performance of question answering systems in a low-resource setup using the cascading of language and task adapters\cite{pfeiffer-etal-2021-adapterfusion,pfeiffer-etal-2020-adapterhub,bapna-firat-2019-simple}. Our work\footnote{Our code and trained models are available at : \url{https://github.com/Bhavik-Ardeshna/Question-Answering-for-Low-Resource-Languages}} contributes by evaluating cross-lingual performance in seven languages - Hindi, Arabic, German, Spanish, English, Vietnamese and Simplified Chinese. Our models are evaluated on the combination of XQuAD\cite{artetxe-etal-2020-cross} and MLQA\cite{lewis-etal-2020-mlqa} datasets which are similar to SQuAD \cite{rajpurkar-etal-2016-squad} . 

To this end, our contributions are as follows:  
\begin{table*}
\centering
\begin{tabular}{llllllll}
\hline
      & \textbf{Hindi} & \textbf{German} & \textbf{Spanish} & \textbf{Arabic} & \textbf{Chinese} & \textbf{Vietnamese} & \textbf{English} \\ \hline
Train & \ 6854             & \ 5707              & \ 6443               & \ 6525             & \ 6327               & \ 6685                  & \ 12780               \\ \hline
Test  & \ 507             & \ 512              & \ 500               & \ 517              & \ 504               & \ 511                  & \ 1148               \\ \hline
\end{tabular}

\caption{Size of the train and test set used in the experiments. The MLQA\cite{lewis-etal-2020-mlqa} test and XQuAD\cite{artetxe-etal-2020-cross} datasets are used for fine-tuning the model and for testing purpose MLQA devset is used for all languages to maintain consistency.}
\label{table:datasetlength}
\end{table*}

\begin{itemize}
\item {We have trained multilingual variants of transformers, namely mBert and XLM-RoBERTa with a QA dataset in seven languages. Both the MLQA and XQuAD datasets contain validation and test sets for the above languages but not the training set. To finetune the model we have combined the test set of XQuAD and MLQA datasets and evaluated the model with the MLQA development dataset as the test dataset. By splitting the dataset in this way we can get train and test data with the considerable length for low-resource languages which helped us to conduct various experiments. Table \ref{table:datasetlength} highlights the size of our train and test set for all the above-mentioned languages.}

\item {We exhaustively analysed the fine-tuned models by evaluating them with the tasks adapter\footnote{Pre-trained task adapters from \url{https://adapterhub.ml/explore/qa/squad1/}} \cite{pfeiffer-etal-2021-adapterfusion,pfeiffer-etal-2020-adapterhub}. We conducted the experiments in two different setups, Houlsby\cite{pmlr-v97-houlsby19a} and Pfeiffer\cite{pfeiffer-etal-2021-adapterfusion,pfeiffer-etal-2020-mad}. These two setups enabled us to compare our language model variants with their multilingual counterparts and understand the different factors that lead to better results on the downstream tasks.}

\item {We have also attempted a series of two different experiments by stacking language adapters and task adapter\footnote{Pre-Trained Language Adapters from \url{https://adapterhub.ml/explore/text_lang/}} in different ways. We first analyze the fine-tuned model by stacking language-specific adapter with the XLM-RoBERTa\textsubscript{\textbf{\emph{base}}} \footnote{XLM-RoBERTa\textsubscript{\textbf{\emph{base}}} \url{https://huggingface.co/deepset/roberta-base-squad2}}. After fine-tuning the language-specific adapter we augment the task-specific adapter upon the previously fine-tuned language adapter. We analyze both the experiments separately and conclude that multiple adapters with the transformer-based model perform notably better.}

\item {Due to limited training, the transfer-learning performance of the transformer is poor on the low-resource languages as well as on the languages unseen during the pretraining\cite{kakwani-etal-2020-indicnlpsuite}. The multi-task adapter (MAD-X) \cite{pfeiffer-etal-2020-mad} outperforms the state-of-the-art models in cross-lingual transfer across a representative set of typologically diverse languages on question answering. To avoid the training of model individually for multiple languages while maintaining the performance, we used cross-lingual transfer by switching heads of language adapter from the source language to the target language.}
\end{itemize}



\section{Proposed Approach}

\begin{figure*}[ht]
    \centering
        \centering\includegraphics[width=2\columnwidth]{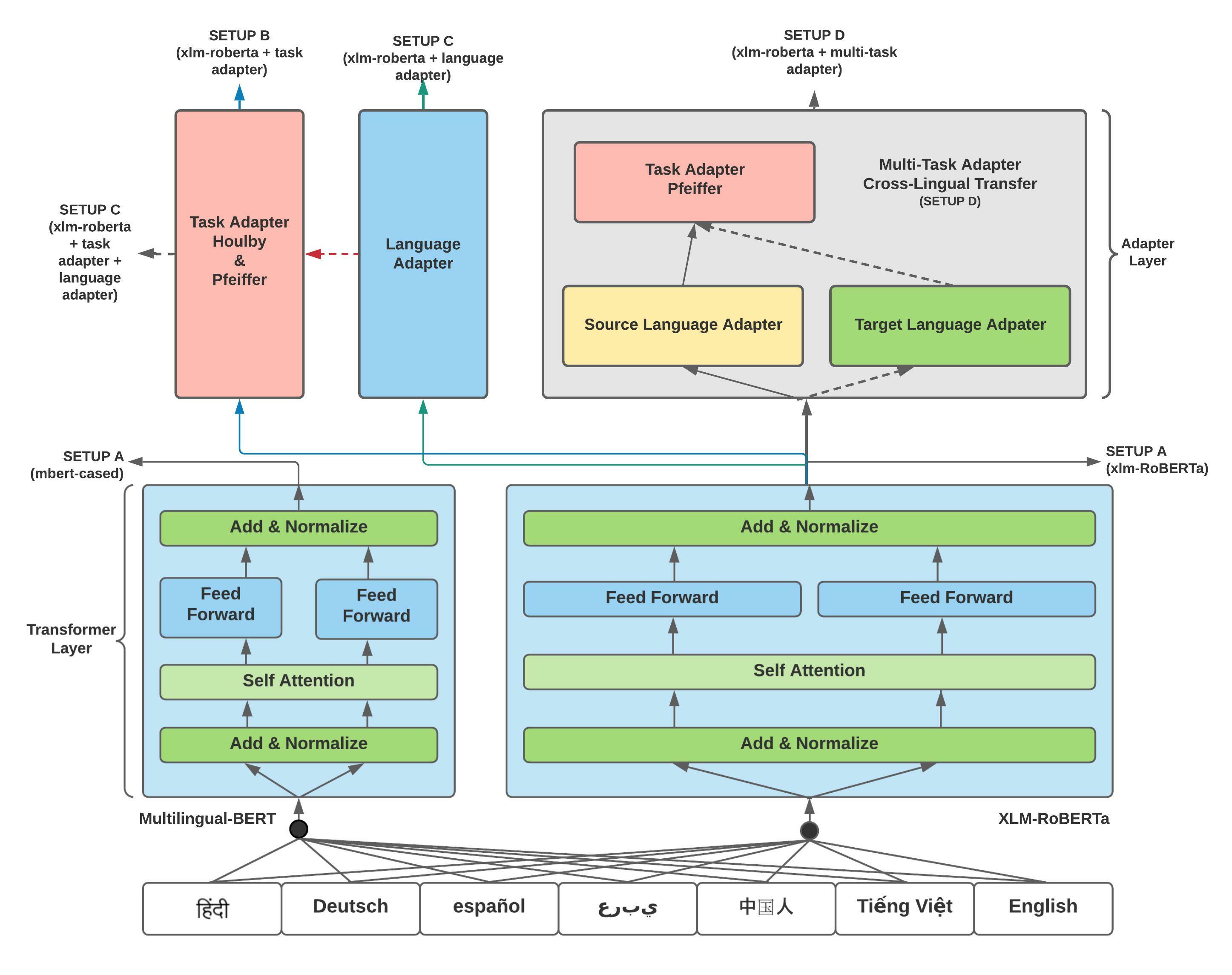}
        \caption{Experimental architecture- SETUP A: mBERT and XLM-R for QA, SETUP B: XLM-R with task adapters setup, SETUP C: XLM-R with language and language + task adapter and SETUP D: MAD-X setup for XLM-R}
        \label{fig:example12}
\end{figure*}
In this section we describe our approach of training the task adapter and the language adapters in 4 different setup.
\subsection{Cross-Lingual Tuning of Task Adapter and Language Adapters}

\textbf{Task-Specific Cross-Lingual Transfer:}  We have used two different configurations for fine-tuning the task-specific adapter for cross-lingual transfer in low-resource languages \cite{pfeiffer-etal-2021-adapterfusion,pmlr-v97-houlsby19a}. We have fine-tuned XLM-RoBERTa\textsubscript{\textbf{\emph{base}}} for multiple languages with the question answering corpora. We calculated the F1-Score, Exact Match, Jaccard \footnote{Jaccard score \url{https://en.wikipedia.org/wiki/Jaccard_index}} , and WER (Word Error Rate)\cite{inproceedings} \footnote{WER score \url{https://en.wikipedia.org/wiki/Word_error_rate}} for the test dataset.

\textbf{Adapting Cross-Lingual learning using Language-Specific Model:} We used the language adapter trained using unlabelled data on MLM objective. It makes the pretrained multilingual model more suitable for the specific language with its improved language understanding. We perform the downstream task by stacking specific language adapter with the XLM-RoBERTa\textsubscript{\textbf{\emph{base}}} and used recent efficient adapter architecture proposed by pfeiffer et al. \cite{pfeiffer-etal-2021-adapterfusion}.

After fine-tuning task-specific adapter and language-specific adapters individually with the different low-resource languages, we observed that by stacking task adapter and language adapters together with the transformer model the performance improved significantly. For each language available in MLQA, we fine-tuned a task adapter using a corresponding question answering dataset.

\subsection{Multi-Task Adapter for Cross-Lingual Transfer}
The adapter-based MAD-X framework \cite{pfeiffer-etal-2020-mad} enables learning language-specific and task-specific transformations in a modular and parameter-efficient way. Our method of using MAD-X is as follows:
\begin{enumerate}
\setlength{\itemsep}{0pt}
  \item We have used pre-trained language adapters\footnote{from \url{https://adapterhub.ml/}} for the source and target language on a language modeling task.
  \item Train a task adapter on the target task dataset. This task adapter is stacked upon the previously trained language adapter. During this step, only the weights of the task adapter are updated.
  \item Next, in zero-shot cross-lingual transfer step, we replaced the source language adapter with the target language adapter while keeping the stacked task adapter.
\end{enumerate}

\section{Experimental setups}
\setlength{\itemsep}{1pt}
\setlength{\parskip}{0pt}
\setlength{\parsep}{1pt}
We have performed 4 different analysis as represented in Figure \ref{fig:example12}. Details of all 4 setups are shown below:

\begin{table*}
\centering
\scalebox{0.75}{
\begin{tabular}{cccccccc}
\hline
                    & \textbf{Hindi} & \textbf{German} & \textbf{Spanish} & \textbf{Arabic} & \textbf{Chinese} & \textbf{Vietnamese} & \textbf{English} \\ \hline
mBERT              & 56.25 / 39.45  & 52.99 / 38.09    & 59.89 / 40.4     & 51.28 / 31.33    & 41.86 / 41.07     & 59.52 / 39.73        & 77.86 / 63.85     \\ \hline
XLM-RoBERTa\textsubscript{\emph{base}}  & 64.49 / 48.32   & 60.74 / 45.31   & 68.99 / 47.6     & 58.07 / 39.65   & 45.37 / 44.24     & 68.19 / 48.53  & 81.29 / 68.64     \\ \hline
XLM-RoBERTa\textsubscript{\emph{large}} & 73.37 / 56.02   & 70.57 / 53.32    & 76.32 / 54.2     & 67.15 / 47.78    & 49.94 / 49.21     & 73.78 / 54.21        & 85.98 / 74.39     \\ \hline
\end{tabular}}
\caption{F1 score and Exact Match on the test set for the Setup A on multilingual-BERT and XLM-RoBERTa.}
\label{table:setupA}
\end{table*}

\begin{table*}
\centering
\scalebox{0.7}{
\begin{tabular}{cccccccc}
\hline
         & \textbf{Hindi} & \textbf{German} & \textbf{Spanish} & \textbf{Arabic} & \textbf{Chinese} & \textbf{Vietnamese} & \textbf{English} \\ \hline
Task Adapter (\emph{Houlby})   & \ 64.12 / 47.73          & \textbf{60.95 / 44.53 }          & \ 68.48 / 46.6            & \textbf{\ 58.13 / 38.49}           & \ \textbf{44.38 / 43.25}            & \ 68.39 / 48.34               & \ 80.86 / 68.29            \\ \hline
Task Adapter (\emph{Pfeiffer}) & \ \textbf{65.7 / 49.9}          & \ 60.53 / 44.14           & \ \textbf{69.09 / 48 }           & \ 55.97 / 37.14           & \ 44.05 / 43.05            & \ \textbf{68.46 / 48.53 }              & \ \textbf{81.23 / 68.64   }         \\ \hline
\end{tabular}}
\caption{F1 score and Exact Match for the xlm-roberta with Task Adapter (Setup B). We bold the best results.}
\label{table:setupB}
\end{table*}

\begin{table*}
\centering
\scalebox{0.7}{
\begin{tabular}{cccccccc}
\hline
         & \textbf{Hindi} & \textbf{German} & \textbf{Spanish} & \textbf{Arabic} & \textbf{Chinese} & \textbf{Vietnamese} & \textbf{English} \\ \hline
Language Adapter   & \ \textbf{66.14 / 49.11}          & \ \textbf{61.41 / 45.9}           & \ \textbf{70.25 / 49.6}            & \ 56.84 / 37.52           & \ \textbf{44.82 / 43.85 }           & \ 68.06 / \textbf{49.31}               & \ \textbf{81.43} / 68.64            \\ \hline
Task + Language Adapter & \ 65.39 / 48.72          & \ 61.03 / 45.51           & \ 69.03 / 47.6            & \ \textbf{58.15 / 38.29}           & \ 44.68 / 43.45            & \ \textbf{68.39} / 48.14               & \ 81.31 / \textbf{68.9}        \\ \hline
\end{tabular}}

\caption{F1 score and Exact Match on the test set for the Setup C and We bold the best result in each section.}
\label{table:setupC}
\end{table*}

\begin{table*}
\centering
\scalebox{0.78}{
\begin{tabular}{cccccccc}
\hline & \textbf{Hindi} & \textbf{German} & \textbf{Spanish} & \textbf{Arabic} & \textbf{Chinese} & \textbf{Vietnamese} & \textbf{English} \\ \hline
XLM-RoBERTa\textsubscript{\emph{base}}               & 59.1 / 76.9 & 51 / 94.9  & 53 / \textbf{74.4}  & \textbf{50} / \textbf{92.7}   & \textbf{44.8} / 60.2 & 57.2 / 81.6 & 73.6 / 49.3 \\ \hline
Task Adapter (\emph{Pfeiffer})    & 58.2 / 84.7 & 51 / \textbf{93}   & 52.2 / 88.9 & 49 / 92.8   & 43.9 / \textbf{59.2} & \textbf{57.3} / 81  & 73.4 / 50.7 \\ \hline
Task Adapter (\emph{Houlby}){      }& 59.7 / \textbf{70.6} & 51.1 / 93.4 & 53.1 / 78.9 & 48.5 / 87.6 & 43.6 / 60.1 & 57.1 / 79.9 & 73.6 / 49.7 \\ \hline
Language Adapter           & \textbf{60.4} / 74.7  & \textbf{52.5} / 104.2 & \textbf{54.6} / 75.9 & 49.2 / 93.8 & 44.3 / 59.7 & 56.6 / \textbf{76.4} & \textbf{73.8} / 49.4 \\ \hline
Task $+$ Language Adapter    & 59.5 / 82.8 & 51.6 / 97.4 & 53 / 76.9  & 49.8 / 93.7 & 44.1 / 59.4 & 57.2 / 85.2 & 73.7 / \textbf{46.5} \\ \hline
MAD-X (Multi-Task Adapter) & 59.7 / 72.6 & 48.6 / 95.5 & 50.3 / 88.7 & 42.9 / 107  & 42.4 / 60.9 & 53.7 / 88.4 & -  \\ \hline
\end{tabular}}

\caption{Jaccard and Word error rate (WER) on the test set for Setup A, B, C, and D}
\label{table:jaccard}
\end{table*}

\begin{table}
\centering
\scalebox{0.7}{
\begin{tabular}{cc}
\hline
 & \textbf{\begin{tabular}[c]{@{}c@{}}Multi-Task Adapter\\ (Task + Source Language + Target Language)\end{tabular}} \\ \hline
Hindi      & \ 65.24 / 48.91 \\ \hline
German     & \ 60.42 / 43.35  \\ \hline
Spanish    & \ 65.82 / 44.2 \\ \hline
Arabic     & \ 50.12 / 31.33 \\ \hline
Chinese    & \ 42.87 / 41.86 \\ \hline
Vietnamese & \ 64.48 / 44.22  \\ \hline
English    & \ - \\ \hline
\end{tabular}}
\caption{F1 score and Exact Match on the test set for the Setup D of Multi-Task adapter}
\label{table:setupD}
\end{table}

\subsection{Setup A}
Here, we evaluated mBERT, XLM-Roberta\textsubscript{\textbf{\emph{base}}} and XLM-Roberta\textsubscript{\textbf{\emph{large}}} models on downstream tasks with the training dataset, which is specific to the individual language variant. The EM and F1 score for all languages are shown in Table \ref{table:setupA}. 

Here, the interpretation of the matrix is F1/EM and it is same for rest of the Setups. For Example, in Table \ref{table:setupA} first entry 56.25/39.45 indicates, for the Hindi test set, the F1score$=$56.25 and EM$=$39.45 is achieved using mBERT transformer model.

\subsection{Setup B}

After fine-tuning the transformer model, We have evaluated XLM-RoBERTa\textsubscript{\textbf{\emph{base}}} with the task-specific adapter on downstream tasks under two training settings: Houlby\cite{pmlr-v97-houlsby19a} and Pfeiffer\cite{pfeiffer-etal-2021-adapterfusion}. While fine-tuning, the weights of only the task adapter get updated and the model weights are kept unchanged. This setup enables the scalable sharing of the task adapter model particularly in low-resource scenarios. Pre-trained task-specific adapters: Houlby\footnote{Available at \url{https://adapterhub.ml/adapters/ukp/roberta-base_qa_squad1_houlsby/}} and Pfeiffer\footnote{ \url{https://adapterhub.ml/adapters/ukp/roberta-base_qa_squad1_pfeiffer/}}  are taken with predefined conditions. The EM and F1 score for all languages are shown in Table \ref{table:setupB}.

\subsection{Setup C}

The language adapters are used to learn language-specific transformations \cite{pfeiffer-etal-2020-mad}. After being trained on a language modeling task, a language adapter can be stacked before a task adapter for training on a downstream task. To perform zero-shot cross-lingual transfer, one language adapter can be replaced by another. In terms of architecture, language adapters are largely similar to task adapters, except for an additional invertible adapter layer after the embedding layer.

In this setup, we have evaluated each language-specific adapter\footnote{Available at \url{https://adapterhub.ml/explore/text_lang/ }} by stacking it on the XLM-RoBERTa model. In the second phase, we stacked the task-specific adapter and language-specific adapter on the XLM-RoBERTa model. The EM and F1 score for the language adapter and the task + language adapter fusion are shown in Table \ref{table:setupC}.

\subsection{Setup D}

Here, we have cascaded the multi-task adapters\cite{pfeiffer-etal-2020-mad} to leverage the high-resource dataset to improve the performance of the low-resource language. We stacked the fine-tuned task-specific adapter upon the language-specific adapter and XLM-RoBERTa (shown in figure \ref{fig:example12}). After fine-tuning with high resource language, we performed zero-shot cross-lingual transfer by switching the source language adapter with the target language adapters. Our results for multi-task adapters are highlighted in the Table \ref{table:setupD}.

\begin{figure}[!h]
    \centering
        \centering\includegraphics[width=1\columnwidth]{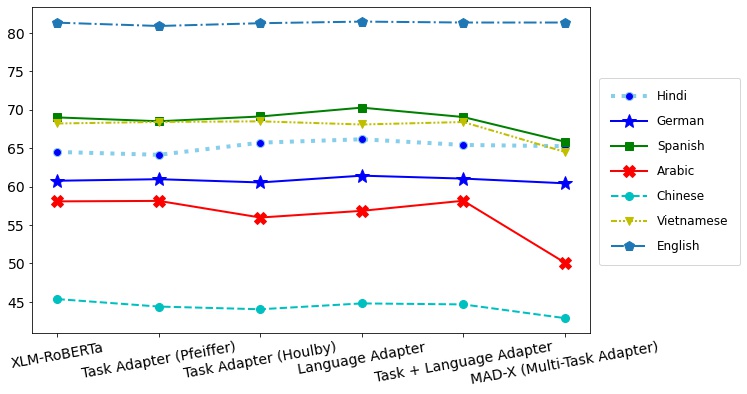}
        \caption{The performance of the different heads. The Y-axis here denotes the F1 score}
        \label{fig:performanceFig}
\end{figure}

Table \ref{table:jaccard} shows Jaccard and WER score for all four setups while the Figure \ref{fig:performanceFig} represents the F1 score of our models on all the languages.

\section{Observations}

To study the impact of the task adapter and the language adapters, we have conducted experiments as shown in Setup B and Setup C. Our observations from Table \ref{table:setupB} and Table \ref{table:setupC} indicates that the trained language adapter (Setup C: language adapters only) improves the performance for Hindi, German, Spanish, Chinese and English languages over the usage of task adapter(Setup B). However, instead of using language adapters only the stack of task and the language adapters lower EM and F1 score for languages other than Arabic.

We have compared two task adapter architectures and noted that the usage of different task adapter architectures have negligible performance impact on majority of the languages. As a result, no clear distinction can be drawn from this observation, which can be used to guide future research.  

High-resource languages that use the Left-to-right(LTR) scripting approach dominate the training of pretrained transformer models. The Arabic language follows Right-to-Left (RTL) scripting style. The general poor performance in the Arabic language could be due to a variation in scripting technique. This also demonstrates that, regardless of the downstream task, the language structure has a significant impact on overall performance. 

The Chinese language has a symbolic language structure and can be written in a variety of forms (right-to-left, or vertically top-to-bottom). The degraded findings in Chinese compared to other low-resource languages are most likely due to the language's writing flexibility.

\section*{Acknowledgments}
The PARAM Shavak HPC computer facility is used for some of our experiments. We are grateful to the Gujarat Council of Science and Technology (GUJCOST) for providing this facility to the institution so that deep learning studies are being carried out effectively.

\section{Conclusions}
We have investigated the efficacy of cascading adapters with transformer models to leverage high-resource language to improve the performance of low-resource languages on the question answering task. We trained four variants of adapter combinations for - Hindi, Arabic, German, Spanish, English, Vietnamese, and Simplified Chinese languages. We demonstrated that by using the transformer model with the multi-task adapters, the performance can be improved for the downstream task. Our results and analysis provide new insights into the generalization abilities of multilingual models for cross-lingual transfer on question answering tasks.

\bibliography{anthology,acl2020}

\begin{thebibliography}{20}
\expandafter\ifx\csname natexlab\endcsname\relax\def\natexlab#1{#1}\fi

\bibitem[{Artetxe et~al.(2020)Artetxe, Ruder, and
  Yogatama}]{artetxe-etal-2020-cross}
Mikel Artetxe, Sebastian Ruder, and Dani Yogatama. 2020.
\newblock \href {https://doi.org/10.18653/v1/2020.acl-main.421} {On the
  cross-lingual transferability of monolingual representations}.
\newblock In \emph{Proceedings of the 58th Annual Meeting of the Association
  for Computational Linguistics}, pages 4623--4637, Online. Association for
  Computational Linguistics.

\bibitem[{Bapna and Firat(2019)}]{bapna-firat-2019-simple}
Ankur Bapna and Orhan Firat. 2019.
\newblock \href {https://doi.org/10.18653/v1/D19-1165} {Simple, scalable
  adaptation for neural machine translation}.
\newblock In \emph{Proceedings of the 2019 Conference on Empirical Methods in
  Natural Language Processing and the 9th International Joint Conference on
  Natural Language Processing (EMNLP-IJCNLP)}, pages 1538--1548, Hong Kong,
  China. Association for Computational Linguistics.

\bibitem[{Baxi et~al.(2015)Baxi, Patel, and
  Bhatt}]{baxi-etal-2015-morphological}
Jatayu Baxi, Pooja Patel, and Brijesh Bhatt. 2015.
\newblock \href {https://www.aclweb.org/anthology/W15-5927} {Morphological
  analyzer for {G}ujarati using paradigm based approach with knowledge based
  and statistical methods}.
\newblock In \emph{Proceedings of the 12th International Conference on Natural
  Language Processing}, pages 178--182, Trivandrum, India. NLP Association of
  India.

\bibitem[{Conneau et~al.(2020)Conneau, Khandelwal, Goyal, Chaudhary, Wenzek,
  Guzm{\'a}n, Grave, Ott, Zettlemoyer, and
  Stoyanov}]{conneau-etal-2020-unsupervised}
Alexis Conneau, Kartikay Khandelwal, Naman Goyal, Vishrav Chaudhary, Guillaume
  Wenzek, Francisco Guzm{\'a}n, Edouard Grave, Myle Ott, Luke Zettlemoyer, and
  Veselin Stoyanov. 2020.
\newblock \href {https://doi.org/10.18653/v1/2020.acl-main.747} {Unsupervised
  cross-lingual representation learning at scale}.
\newblock In \emph{Proceedings of the 58th Annual Meeting of the Association
  for Computational Linguistics}, pages 8440--8451, Online. Association for
  Computational Linguistics.

\bibitem[{Delobelle et~al.(2020)Delobelle, Winters, and
  Berendt}]{delobelle-etal-2020-robbert}
Pieter Delobelle, Thomas Winters, and Bettina Berendt. 2020.
\newblock \href {https://doi.org/10.18653/v1/2020.findings-emnlp.292}
  {{R}ob{BERT}: a {D}utch {R}o{BERT}a-based {L}anguage {M}odel}.
\newblock In \emph{Findings of the Association for Computational Linguistics:
  EMNLP 2020}, pages 3255--3265, Online. Association for Computational
  Linguistics.

\bibitem[{Devlin et~al.(2019)Devlin, Chang, Lee, and
  Toutanova}]{devlin-etal-2019-bert}
Jacob Devlin, Ming-Wei Chang, Kenton Lee, and Kristina Toutanova. 2019.
\newblock \href {https://doi.org/10.18653/v1/N19-1423} {{BERT}: Pre-training of
  deep bidirectional transformers for language understanding}.
\newblock In \emph{Proceedings of the 2019 Conference of the North {A}merican
  Chapter of the Association for Computational Linguistics: Human Language
  Technologies, Volume 1 (Long and Short Papers)}, pages 4171--4186,
  Minneapolis, Minnesota. Association for Computational Linguistics.

\bibitem[{Houlsby et~al.(2019)Houlsby, Giurgiu, Jastrzebski, Morrone,
  De~Laroussilhe, Gesmundo, Attariyan, and Gelly}]{pmlr-v97-houlsby19a}
Neil Houlsby, Andrei Giurgiu, Stanislaw Jastrzebski, Bruna Morrone, Quentin
  De~Laroussilhe, Andrea Gesmundo, Mona Attariyan, and Sylvain Gelly. 2019.
\newblock \href {https://proceedings.mlr.press/v97/houlsby19a.html}
  {Parameter-efficient transfer learning for {NLP}}.
\newblock In \emph{Proceedings of the 36th International Conference on Machine
  Learning}, volume~97 of \emph{Proceedings of Machine Learning Research},
  pages 2790--2799. PMLR.

\bibitem[{Kakwani et~al.(2020)Kakwani, Kunchukuttan, Golla, N.C.,
  Bhattacharyya, Khapra, and Kumar}]{kakwani-etal-2020-indicnlpsuite}
Divyanshu Kakwani, Anoop Kunchukuttan, Satish Golla, Gokul N.C., Avik
  Bhattacharyya, Mitesh~M. Khapra, and Pratyush Kumar. 2020.
\newblock \href {https://doi.org/10.18653/v1/2020.findings-emnlp.445}
  {{I}ndic{NLPS}uite: Monolingual corpora, evaluation benchmarks and
  pre-trained multilingual language models for {I}ndian languages}.
\newblock In \emph{Findings of the Association for Computational Linguistics:
  EMNLP 2020}, pages 4948--4961, Online. Association for Computational
  Linguistics.

\bibitem[{Lewis et~al.(2020)Lewis, Oguz, Rinott, Riedel, and
  Schwenk}]{lewis-etal-2020-mlqa}
Patrick Lewis, Barlas Oguz, Ruty Rinott, Sebastian Riedel, and Holger Schwenk.
  2020.
\newblock \href {https://doi.org/10.18653/v1/2020.acl-main.653} {{MLQA}:
  Evaluating cross-lingual extractive question answering}.
\newblock In \emph{Proceedings of the 58th Annual Meeting of the Association
  for Computational Linguistics}, pages 7315--7330, Online. Association for
  Computational Linguistics.

\bibitem[{Murthy et~al.(2019)Murthy, Kunchukuttan, and
  Bhattacharyya}]{murthy-etal-2019-addressing}
Rudra Murthy, Anoop Kunchukuttan, and Pushpak Bhattacharyya. 2019.
\newblock \href {https://doi.org/10.18653/v1/N19-1387} {Addressing word-order
  divergence in multilingual neural machine translation for extremely low
  resource languages}.
\newblock In \emph{Proceedings of the 2019 Conference of the North {A}merican
  Chapter of the Association for Computational Linguistics: Human Language
  Technologies, Volume 1 (Long and Short Papers)}, pages 3868--3873,
  Minneapolis, Minnesota. Association for Computational Linguistics.

\bibitem[{Pandya and Bhatt(2021)}]{pandya2021question}
HA~Pandya and BS~Bhatt. 2021.
\newblock \href {https://doi.org/10.37896/jxu15.4/014} {Question answering
  survey: Directions, challenges, datasets, evaluation matrices}.
\newblock \emph{Journal of Xidian University}, 15(4):152--168.

\bibitem[{Park et~al.(2008)Park, Patwardhan, Visweswariah, and
  Gates}]{inproceedings}
Youngja Park, Siddharth Patwardhan, Karthik Visweswariah, and Stephen Gates.
  2008.
\newblock \href {https://doi.org/10.21437/Interspeech.2008-537} {An empirical
  analysis of word error rate and keyword error rate}.
\newblock pages 2070--2073.

\bibitem[{Pfeiffer et~al.(2021)Pfeiffer, Kamath, R{\"u}ckl{\'e}, Cho, and
  Gurevych}]{pfeiffer-etal-2021-adapterfusion}
Jonas Pfeiffer, Aishwarya Kamath, Andreas R{\"u}ckl{\'e}, Kyunghyun Cho, and
  Iryna Gurevych. 2021.
\newblock \href {https://aclanthology.org/2021.eacl-main.39}
  {{A}dapter{F}usion: Non-destructive task composition for transfer learning}.
\newblock In \emph{Proceedings of the 16th Conference of the European Chapter
  of the Association for Computational Linguistics: Main Volume}, pages
  487--503, Online. Association for Computational Linguistics.

\bibitem[{Pfeiffer et~al.(2020{\natexlab{a}})Pfeiffer, R{\"u}ckl{\'e}, Poth,
  Kamath, Vuli{\'c}, Ruder, Cho, and Gurevych}]{pfeiffer-etal-2020-adapterhub}
Jonas Pfeiffer, Andreas R{\"u}ckl{\'e}, Clifton Poth, Aishwarya Kamath, Ivan
  Vuli{\'c}, Sebastian Ruder, Kyunghyun Cho, and Iryna Gurevych.
  2020{\natexlab{a}}.
\newblock \href {https://doi.org/10.18653/v1/2020.emnlp-demos.7}
  {{A}dapter{H}ub: A framework for adapting transformers}.
\newblock In \emph{Proceedings of the 2020 Conference on Empirical Methods in
  Natural Language Processing: System Demonstrations}, pages 46--54, Online.
  Association for Computational Linguistics.

\bibitem[{Pfeiffer et~al.(2020{\natexlab{b}})Pfeiffer, Vuli{\'c}, Gurevych, and
  Ruder}]{pfeiffer-etal-2020-mad}
Jonas Pfeiffer, Ivan Vuli{\'c}, Iryna Gurevych, and Sebastian Ruder.
  2020{\natexlab{b}}.
\newblock \href {https://doi.org/10.18653/v1/2020.emnlp-main.617} {{MAD-X}:
  {A}n {A}dapter-{B}ased {F}ramework for {M}ulti-{T}ask {C}ross-{L}ingual
  {T}ransfer}.
\newblock In \emph{Proceedings of the 2020 Conference on Empirical Methods in
  Natural Language Processing (EMNLP)}, pages 7654--7673, Online. Association
  for Computational Linguistics.

\bibitem[{Pires et~al.(2019)Pires, Schlinger, and
  Garrette}]{pires-etal-2019-multilingual}
Telmo Pires, Eva Schlinger, and Dan Garrette. 2019.
\newblock \href {https://doi.org/10.18653/v1/P19-1493} {How multilingual is
  multilingual {BERT}?}
\newblock In \emph{Proceedings of the 57th Annual Meeting of the Association
  for Computational Linguistics}, pages 4996--5001, Florence, Italy.
  Association for Computational Linguistics.

\bibitem[{Raffel et~al.(2020)Raffel, Shazeer, Roberts, Lee, Narang, Matena,
  Zhou, Li, and Liu}]{JMLR:v21:20-074}
Colin Raffel, Noam Shazeer, Adam Roberts, Katherine Lee, Sharan Narang, Michael
  Matena, Yanqi Zhou, Wei Li, and Peter~J. Liu. 2020.
\newblock \href {http://jmlr.org/papers/v21/20-074.html} {Exploring the limits
  of transfer learning with a unified text-to-text transformer}.
\newblock \emph{Journal of Machine Learning Research}, 21(140):1--67.

\bibitem[{Rajpurkar et~al.(2016)Rajpurkar, Zhang, Lopyrev, and
  Liang}]{rajpurkar-etal-2016-squad}
Pranav Rajpurkar, Jian Zhang, Konstantin Lopyrev, and Percy Liang. 2016.
\newblock \href {https://doi.org/10.18653/v1/D16-1264} {{SQ}u{AD}: 100,000+
  questions for machine comprehension of text}.
\newblock In \emph{Proceedings of the 2016 Conference on Empirical Methods in
  Natural Language Processing}, pages 2383--2392, Austin, Texas. Association
  for Computational Linguistics.

\bibitem[{Saha et~al.(2021)Saha, Gupta, Saha, and
  Bhattacharyya}]{10.1145/3461763}
Tulika Saha, Dhawal Gupta, Sriparna Saha, and Pushpak Bhattacharyya. 2021.
\newblock \href {https://doi.org/10.1145/3461763} {A unified dialogue
  management strategy for multi-intent dialogue conversations in multiple
  languages}.
\newblock \emph{ACM Trans. Asian Low-Resour. Lang. Inf. Process.}, 20(6).

\bibitem[{Yang et~al.(2019)Yang, Dai, Yang, Carbonell, Salakhutdinov, and
  Le}]{NEURIPS2019_dc6a7e65}
Zhilin Yang, Zihang Dai, Yiming Yang, Jaime Carbonell, Russ~R Salakhutdinov,
  and Quoc~V Le. 2019.
\newblock \href
  {https://proceedings.neurips.cc/paper/2019/file/dc6a7e655d7e5840e66733e9ee67cc69-Paper.pdf}
  {Xlnet: Generalized autoregressive pretraining for language understanding}.
\newblock In \emph{Advances in Neural Information Processing Systems},
  volume~32. Curran Associates, Inc.

\end{thebibliography}
\bibliographystyle{acl_natbib}

\end{document}